\documentclass{article}
\usepackage{ijcai19}
\usepackage{times}
\usepackage{soul}
\usepackage{url}
\usepackage[hidelinks]{hyperref}
\usepackage[utf8]{inputenc}
\usepackage[small]{caption}
\usepackage{graphicx}
\usepackage{amsmath,amsfonts}
\usepackage{booktabs}
\usepackage{multirow}

\usepackage{float}

\usepackage[ruled,vlined]{algorithm2e} % algorithms. 

\usepackage{xcolor}

\newcommand{\citep}[1]{\cite{#1}}
\newcommand{\citet}[1]{\citeauthor{#1} [\citeyear{#1}]}

\title{Safe Contextual Bayesian Optimization for Sustainable Room Temperature PID Control Tuning}

\author{
Marcello Fiducioso$^{1, 2}$\and
Sebastian Curi$^1$\footnote{Contact Author}\and 
Benedikt Schumacher$^2$\and \\
Markus Gwerder$^2$\And
Andreas Krause$^1$\\
\affiliations
$^1$ Department of Computer Science, ETH Zurich, Switzerland\\
$^2$ Building Technologies Division, Siemens Switzerland Ltd.\\
\emails
marcello.fiducioso@gmail.com,
sebastian.curi@inf.ethz.ch,
benedikt.schumacher@siemens.com,
markus.gwerder@siemens.com,
krausea@ethz.ch
}

\begin{document}

\maketitle

\begin{abstract}
We tune one of the most common heating, ventilation, and air conditioning (HVAC) control loops, namely the temperature control of a room. For economical and environmental reasons, it is of prime importance to optimize the performance of this system. Buildings account from 20 to 40\% of a country energy consumption, and almost 50\% of it comes from HVAC systems. Scenario projections predict a 30\% decrease in heating consumption by 2050 due to efficiency increase. Advanced control techniques can improve performance; however, the proportional-integral-derivative (PID) control is typically used due to its simplicity and overall performance. We use Safe Contextual Bayesian Optimization to optimize the PID parameters without human intervention. We reduce costs by 32\% compared to the current PID controller setting while assuring safety and comfort to people in the room. The results of this work have an immediate impact on the room control loop performances and its related commissioning costs. Furthermore, this successful attempt paves the way for further use at different levels of HVAC systems, with promising energy, operational, and commissioning costs savings, and it is a practical demonstration of the positive effects that Artificial Intelligence can have on environmental sustainability. 
\end{abstract}

\section{Introduction}
Many countries have introduced policies to increase energy savings. Heating, ventilation, and air conditioning (HVAC) systems are the major components of Building energy consumption, which in turn constitute between 20\% to 40\% yearly energy costs\footnote{https://www.eia.gov/outlooks/aeo/} \citep{perez2008review}. Room temperature control loops are one of the most common control loops in heating, ventilation, and air conditioning (HVAC) systems. In these loops, the heater is a hot water radiator or a floor heating system, and the room temperature is controlled to satisfy operational constraints, such as pleasant temperature and low energy consumption. Adaptive control techniques are a promising path to achieve these goals and increase energy savings as proven by the DeepMind Data Center cooling project \citep{evans2016deepmind}. 

The proportional-integral-derivative (PID) controller dominates the market; over 97 \% of controllers in industrial applications have the PID architecture \citep{desborough2002increasing}. Its low cost, ease of use, and proven performance makes it the predominant choice for industrial control loops. Despite its advantages, this architecture often produces sub-optimal performance. Tuning is an expensive, labor-intensive trial-and-error task. Thus, only a few of real-life loops are tuned to achieve optimal performance \citep{odwyer2006handbook}. 

A prominent PID tuning technique is the Step Response Ziegler-Nichols' rule \citep{ziegler1942optimum}, in which an engineer applies an open loop input step and according to the measured response, the PID gains are selected to optimize a performance criterion. This method is entirely model-free but assumes that the system is linear. The performance criteria are disturbance rejection, rising time, or overshoot. If the engineer has a low-order linear model of the system, she can skip the initial phase as she can calculate the parameters in closed form. However, the models are never exact, and the \emph{ad hoc} tuning is trial-and-error around the initial guess. 

The main drawbacks of HVAC PID control with manual tuning are human effort and related commissioning costs. Further, this controller has no adaptation capabilities for changes in buildings. To overcome these limitations, the control community developed linear quadratic regulators (LQR) in the 70s \citep{kwakernaak1972linear}, optimal design $\mathcal{H}_2$ or $\mathcal{H}_\infty$ \citep{zhou1996robust}, and more recently model predictive control (MPC) \citep{garcia1989model}. These methods are model-based, and HVAC system models are very complex and not well understood \citep{oldewurtel2010energy,afram2014theory}. Thus, it is not clear how to apply such advanced control strategies. 

Room temperature control exhibits both complex dynamics and disturbances, and optimizing performance is critical for economic and environmental reasons. Sometimes, an engineer may be sent in situ to tune the controller before it gets deployed. Often, however, off-the-shelf controller settings are used that give adequate performance for the typical range of room temperature control loops but are far from optimal.

In this work, we use Safe Contextual Bayesian Optimization to learn the best PID parameters, while \emph{provably} keeping safe conditions during the optimization procedure for different outside air temperature scenarios. For the reasons exposed before, this method is of prime industrial interest. As time constants are slow, a typical interaction with the system takes one day. Therefore, other data-intensive techniques based on Deep Learning may struggle in such systems. Bayesian Optimization, on the other hand, is \emph{data efficient} as it actively queries points to trade-off exploration and exploitation, without the need to learn an explicit model of the plant dynamics. 

\section{Related Work}
In a recent review of HVAC systems control, \citet{afram2014theory} summarize the documented control strategies and show that the most prominent in the industry are PID, while in research are MPC. In this review, they compare data-driven methods such as artificial neural networks controllers, but state that ``industry is usually reluctant to adopt and use a black-box approach.'' The authors do include two data-driven approaches used in real life \citep{seem1998new,bi2000advanced}. These use model-based approaches: the algorithm fits a first-order model with a delay from data, and with this idealized model, it uses the Ziegler-Nichols rule to select optimal gains. The review paper thus focuses on conventional control strategies and points out MPC as the one strategy that performs best as it can adapt to different operating conditions through weather forecasts, such as in \citep{oldewurtel2010energy}, and it can use learning strategies for the model online, such as in \citep{lixing2010support}. Nevertheless, the MPC algorithm has to solve an optimization problem that is a surrogate of the real criteria and \emph{adapts} to the data by the receding horizon approach, i.e., it proposes a new control input when the current state does not match the prediction. Furthermore, tuning MPC controllers is harder than tuning PID in HVAC systems as it heavily relies on a model, which makes the industry reluctant to adopt this technology. 

Bayesian Optimization is an active learning approach to learn and optimize directly from data a cost function \citep{snoek2012practical}. In this work, we use GP-LCB as the acquisition function criterion \citep{srinivas2009gaussian}, in which the cost function is learned using a Gaussian Process (GP), and the next point is selected in order to maximize the cost lower confidence bound (LCB). \citet{krause2011contextual} included context by augmenting the GP kernel with an extra variable, while \citet{sui2015safe} and \citet{berkenkamp2016bayesian} included safety constraints which restrict the exploration phase. Safety is critical for deployment when people interact with the system. However, the exploration phase with this method might be slow. Another approach that includes safe exploration was presented by \citet{gelbart2014bayesian}. They use an indicator function in the acquisition function to force exploration only in the safe regions. The latter method is faster than the former, but it can get trapped in local optima. 

Even though Bayesian Optimization has not been explicitly used in building control, it has many success stories in controller tuning. For example, \citet{calandra2014experimental} used Bayesian Optimization to tune the gait parameters of a Bipedal walker, showing that GP-LCB outperforms all the other acquisition function approaches. \citet{marco2017virtual} used Bayesian Optimization to tune process and measurement noise covariances jointly and, through an LQR approach, they learn the gains of a state feedback controller. \citet{berkenkamp2016bayesian} optimized the PD parameters of a quadrotor controller and tested it on the vehicle for reference position and circular trajectories tracking. Finally, \citet{abdelrahman2016bayesian} maximized the energy generated over time by a photovoltaic power plant by optimizing the operating voltages. 

\paragraph{Our contributions.} In this work, we optimize the PID parameters of a temperature control loop and obtain a 32\% cost reduction compared to the current PID parameters over a heating season. The main contribution of our work is that we explicitly include the performance indexes that the engineer is interested in this application as opposed to quadratic surrogates that allow closed-form solutions. Furthermore, we can find the optimal controller parameters through safe-active exploration, without any human intervention, and with minimal modification of the existing controller configurations. 

\section{Dynamical System Model}
In room temperature control, a controller regulates with a valve the amount of hot water that passes through a radiator in order to keep the temperature close to a user-defined set-point. The controller parameters are selected to trade-off the energy consumption, the expected lifetime of a valve, the time it takes for the room to reach the desired temperature, and minimize the temperature variations after that. 

\begin{figure}[htpb]
	\begin{center}
	\includegraphics[width=0.4\textwidth]{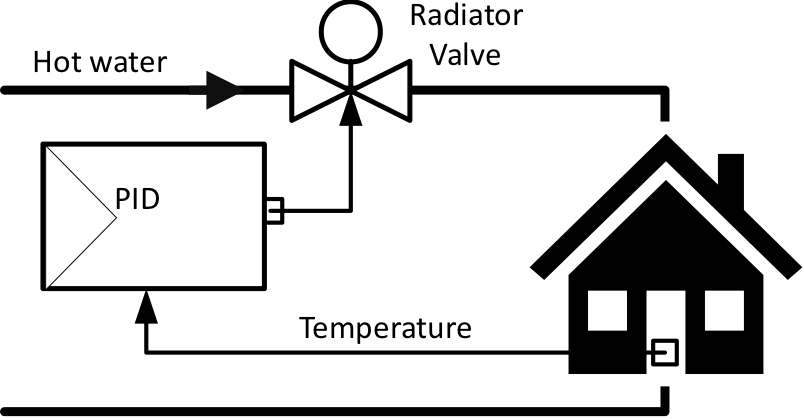}
	\caption{Simplified plant schematic. The controller measures the room temperature and moves a valve that limits the flow of hot water towards the room. The room is equipped with a radiator that transfers the heat from the water to the room.} 
	\label{fig:schema} 
	\end{center}
\end{figure}

Figure~\ref{fig:schema} depicts the piping along with the most important blocks in the loop. To design controllers, usually, a first-order model is used to model the room as follows \citep{deng2009control}:
\begin{equation}
	T_{k+1} = -a T_k + b_u u_k + b_d d_k, \label{eq:first_order}
\end{equation}
where $T_k$ is the temperature at time $k$, $u_k$ is the controlled input, $d_k$ is the disturbance, and $a>0, b_u>0$ and $b_d > 0$ are parameters of the plant. Despite using this extremely simplified model, the resulting controller successfully controls the temperature to the desired set-point but it typically \emph{under-performs}. Furthermore, even if the model in Eq.~\ref{eq:first_order} was correct, different buildings have different parameters $(a, b_u, b_d)$. A robust control design will optimize the performance for the worst case combination of parameters and disturbances, but this is usually too conservative. Finally, the model is too simplistic to be correct. A non-exhaustive list of phenomena that this model does not address is:

\begin{enumerate}
	\item Detailed modeling of the thermal masses dynamics;
	\item Delays due to water transport and valve time constant;
	\item Solar radiation variations;
	\item Outside air temperature;
	\item Thermal gains due to people and equipment;
	\item Interactions with other rooms and control loops;
	\item Disturbances such as open windows and doors.
\end{enumerate}

The effects of the outside air temperature require special attention since it can be easily measured and used to adapt the controller parameters with gain scheduling. In the first place, the heat loss to the environment is inversely proportional to the outside air temperature \cite{bergman2011fundamentals}. Second, the heating generation systems use automatic weather compensation to lower fuel consumption. Heating systems' capacity is designed for the lowest possible outside air temperature at the building site. Since this is a lower bound for the actual temperature, automatic weather compensation lowers the boiler temperature to save energy without decreasing comfort. Consequently, the pumped water temperature fluctuates with outside air temperature, which in turns lowers the \emph{actuation power}. These two competing effects make an \emph{a priori} gain-scheduling technique unfit for this application. Instead, we include the outside air temperature as a context in the Bayesian Optimization algorithm, so it learns from data the effect that this variable has on the control loop. 

The other disturbances are hard or expensive to measure, so we will model their combined effect as zero-mean Gaussian noise with unknown variance but bounded magnitude. 

\section{Problem Statement}
The goal of room temperature control is to (1) drive the temperature of a room to a desired set-point as fast as possible, (2) reduce the room temperature variations thereafter and maintaining comfort, (3) maximize the expected lifetime of the actuators, and (4) use the minimum amount of possible energy. We express these performance criteria as: 
\begin{equation}
	J = \sum_{i=1}^4 w_i J_i, \label{eq:cost function}
\end{equation}
where $J_1$ is the 10-90 \% rise time, $J_2$ is the temperature overshoot, $J_3$ is the $l$-2 norm of the controller output derivative, and $J_4$ is the $l$-2 norm of the controller output. 
The 2-norm of the output derivative cost reduces the valve oscillations, increases its lifetime, and prevents large inputs steps to occur that may result in unpleasant room temperatures. The 2-norm of the controller output measures the energy consumption as a larger norm indicates more hot water passing through the radiator, hence higher heating costs. The cost function in Eq.~\ref{eq:cost function} trades-off the different criteria with user-specified weights $w_i > 0$.

The optimization variables are the proportional and the integral gains, indicated by $a \in \mathcal{A} \subset \mathbb{R}^2$. Usually, the derivative gain is set to zero in this application as the system is inherently stable, and this gain amplifies noise leading to bad performance. The cost $J$ is a function of these parameters, and the goal is to find the parameters that minimize it. 

There are also operational constraints that must be satisfied at all times to ensure safety and feasibility. These constraints are modeled with inequality constraints such as $J_i(a) \leq c_i$. In this work, we use the following constraints. (1), the 10-90 \% rise time has to be smaller than a maximum allowed time to ensure that the temperature reaches the comfort zone when people enter the building, i.e., if this constraint is met and the system is turned on at least this time before people enter the building, the temperature constraints will be satisfied. (2), the overshoot has to be smaller than a maximum temperature deviation to ensure that the temperature stays in a comfortable range. And (3), the $l$-2 norm of the derivative of controller output has to be smaller than a maximum norm. This last constraint serves two purposes: first, it controls the life expectancy of the valve actuators; second, it limits the oscillation frequency around the temperature set-point which affects the comfort of people inside the building. We will provide specific parameters in the experimental section. 

Finally, we also include the outside air temperature as a context for the optimization problem $z \in \mathcal{Z} \subset \mathbb{R}$. This context modifies the cost and constraint functions but does not modify the weights used to trade-off the different costs nor the inequality parameters. Furthermore, the context is not an optimization variable. 

With all these into account, the optimization problem to solve for each $z \in \mathcal{Z}$ is:

\begin{equation}
	\begin{aligned}
	& \underset{a \in \mathcal{A}}{\text{minimize}} & J(a; z) & \\
	& \text{subject to} & J_i(a; z) \leq c_i, & \qquad \forall i \in \left\{1, 2,3 \right\}
	\end{aligned} \label{eq:minimization}
\end{equation}

\section{Optimization Algorithm}
To solve the optimization algorithm in Eq.~\ref{eq:minimization} we use a variation of the GP-LCB algorithm \citep{srinivas2009gaussian}, that uses contextual information as in CGP-LCB \citep{krause2011contextual}, and adds multiple constraints \citep{berkenkamp2016bayesian}, yielding our Safe Contextual GP-LCB. In order to do this, we place a Gaussian Processes (GP) prior to each cost function $J_i$. As a first step, we introduce GP regression. Then, we present the Safe Contextual GP-LCB algorithm. 

\subsection{Gaussian Process Regression}
Gaussian Processes (GPs) are a common non-parametric regression technique in machine learning. The goal is to approximate a non-linear map $f(a): \mathcal{A} \to \mathbb{R}$. The main assumption is that the values of $f(\cdot)$ at different locations of $a$ are random variables that have a joint Gaussian distribution. The GP is completely defined through a prior mean function $\mu(a)$ and a covariance function $k(a, a')$. The choice of the covariance function, also known as the kernel, is crucial for performance. It defines the regularity in the correlation of the marginal distributions of $f(a)$ and $f(a')$ and it is a measure of the similarity between the points $a$ and $a'$. 

The GP framework can be used to learn the mapping from controller parameters and outside air temperature to cost function values. Assume we have a set of $t$ observations of the cost function at different controller parameters and contexts, denoted by $\tilde{a} = (a, z)$, i.e., \ $y_t = [\hat{J}_i(\tilde{a}_1), \ldots, \hat{J}_i(\tilde{a}_t)]$. The observations are assumed to be corrupted with zero mean Gaussian noise, $\hat{J}_i(\tilde{a}_n) = J_i(\tilde{a}_n) + \omega_n$, with $\omega_n \sim \mathcal{N}(0, \sigma_\omega^2)$. Conditioned on these observations, the GP estimates the function $J_i(\cdot)$ at a new location $(a; z)$ with a Gaussian distribution with mean and variance:
\begin{align}
	\mu_t(\tilde{a}; J_i) &= k_t(\tilde{a})(K_t+I_t \sigma^2_{\omega})^{-1}y_t \label{eq:mean_posterior} \\
	\sigma_t^2(\tilde{a}; J_i) &= k(\tilde{a}, \tilde{a})-k_t(\tilde{a})(K_t+I_t \sigma^2_{\omega})^{-1}k_t^T(\tilde{a}) \label{eq:cov_posterior}
\end{align}
	
where $I_t \in \mathbb{R}^{t\times t}$ is the identity matrix, $k_t(a;z) = [k(\tilde{a}, \tilde{a}_1), \dots, k(\tilde{a}, \tilde{a}_t)] \in \mathbb{R}^{t} $, and $K_t \in \mathbb{R}^{t\times t}$ is the kernel matrix with entries $[K_t]_{(m,n)}=k(\tilde{a}_m, \tilde{a}_n), m,n \in {1,\dots,t}$. 

The cost function $J$ defined in Eq.~\eqref{eq:cost function} is also a GP as it is a linear combination of GPs and we use closed-form formulas for its mean and variance \citep{rasmussen2004gaussian}.

\subsection{Safe Contextual GP-LCB}
To optimize the problem defined in Eq.~\ref{eq:minimization}, we will use Safe Contextual GP-LCB as defined in Algorithm~\ref{alg:safe contextual ucb}. The algorithm proceeds in rounds: At the beginning of each round, it observes a context. With it, it constructs a safe set of controller parameters (for this context) and chooses an action that minimizes the lower bound of the estimated cost constrained to the safe set. After interacting with the environment, it observes the realization of each of the cost and updates their respective GP models.

The algorithm has two main differences compared to the original GP-LCB algorithm \citep{srinivas2009gaussian}. First it includes a context as in CGP-LCB \citep{krause2011contextual}. In our setting, the context is the outside air temperature at the beginning of a heating day, which affects the water temperature in the heater and the heat transfer to the environment systematically. This allows factoring in this exogenous disturbance in our model without including it as an optimization variable. In the experimental section, we will discuss the effect of this modeling choice in the performance of our algorithm. The other difference is the inclusion of a safe set. This is also done for single constraints in \citep{sui2015safe} and multiple-constraints \citep{berkenkamp2016bayesian}. 

The main theoretical difference between these papers is that we use a different acquisition function to make the exploration faster. Although their acquisition function has regret guarantees for finite-time algorithms, we use the acquisition function of contextual GP-LCB constrained to the safe set. In SafeOPT, the exploration explicitly selects the most uncertain point over all possible optimizers and potential expanders of the safe set. We found that the exploratory phase was too long in our experimental setting and, after a heating season, we had poor performances. The argument for their acquisition function is that it expands the safe set, while Safe-LCB does not. There are pathological situations in which Safe-LCB might be arbitrarily suboptimal. In our experimental setting, we found that this was not one of these situations. Instead, we demonstrate experimentally that we expand the safe set and quickly find good optimizers.

\begin{algorithm}[ht]
\SetAlgoLined
% \Input{\eta}
\KwIn{Domain $\mathcal{A}$ \\ GP prior $k(\cdot, \cdot)$ \\ Initial, safe controller parameters $a_0$ \\ Exploration parameter $\beta_t$ \\ Risk parameter $\epsilon$}
Initialize 	GP with $a_0, \hat{J}(a_0, \cdot)$ and $a_0, \hat{J}_i(a_0, \cdot)$\\ 
\For{$n = 1, \ldots$}{
	Observe context $z_n$; Define $\tilde{a} \equiv (a, z_n)$ \;
	$\mathcal{S}_n \gets \left\{ a \in \mathcal{A} \vert Pr(J_i(\tilde{a}) \leq c_i)\geq 1-\epsilon \quad \forall i \right\}$ \;
	$a_n \gets \arg\min_{a \in S_n} \mu_{n-1}(\tilde{a}; J) - \beta_n \sigma_{n-1}(\tilde{a}; J) $ \;
	Define $\tilde{a}_n \equiv (a_n, z_n)$ \;
	Observe $\hat{J}_i(\tilde{a}_n) \gets J_i(\tilde{a}_n) + \omega_{i,n}$ \;
	Update GP models with $(\tilde{a}_n; \hat{J}_i(\tilde{a}_n))$ \; 
	}
\caption{Safe contextual GP-LCB.} \label{alg:safe contextual ucb} 
\end{algorithm}

\section{Experimental Setup}

\subsection{Real Data Simulation}
In this work, we use a simulator based on \citet{gao07} provided by Siemens Building Technology Division.\footnote{https://new.siemens.com/global/en/products/buildings.html}. 
The model handles thermal gains due to \emph{solar radiation}, \emph{people presence} and \emph{equipment heating}. The heat exchange with the environment varies according to the \emph{outside air temperature}. The room control loop regulates the amount of radiator inlet water. The water temperature control loop uses the automatic weather compensation system, which determines the water temperature according to the outside air temperature. Our goal is to optimize the room control loop without knowledge of the water temperature nor its set-point. 

The building model ensures a realistic simulation of the thermal dynamics based on solar radiation, heat transfer through airflow, and heat transfer through walls and windows using real measurements, taken in the city of Zug, Switzerland, from the $21^{\text{st}}$ of October 2016 to the $14^{\text{th}}$ of March 2017. The obtained results derive from the simulation of a $25 m^2$ room with $4 m^2$ of windows area.

\subsection{Cost and Constraint Function Details}
The algorithm exploratory parameter is fixed to $\beta = 2$ as it proved to have the best performance in our simulations. The costs are all normalized so that in 95\% of the historical data, they are between 0 and 1. In this work, we select all the weights equal to 0.25 to normalize the total cost between 0 and 1. Similarly, we set the parameters $c_i$ so that 97.5\% of each of the historical costs are considered safe. 

\subsection{Gaussian Process Regression Details}
\paragraph{Mean Functions.} 
Normally, GP models have zero mean. To enhance performance, we use Explicit Basis Functions \citep{rasmussen2004gaussian} to predict the mean of the cost function and then use a zero-mean GP on the residual as: 
\begin{equation*}
	J_i(\cdot) = GP(0, k(\cdot, \cdot)) + \boldsymbol{1}^\top \alpha,
\end{equation*}
where $\boldsymbol{1}$ is a vector of all ones and $\alpha$ are coefficients learned via maximum likelihood, and it is a stochastic mean function. Although this is useful for performance, it might be detrimental for safety, as it might consider safe regions not explored yet due to extrapolation of the mean. To this end, we model the cost and constraint with different GPs and use the explicit basis functions on the cost models only. 

\paragraph{Kernel Selection.}
We use the same kernel family for each cost and constraint model as we expect similar smoothness properties for different models. Although the kernels are the same, the hyperparameters are optimized separately.
The GP model predicts each cost as a function of the PID parameters $a \in \mathcal{A} $, and the outside air temperature $z\in \mathcal{Z}$. We chose the multiplication of kernels $k_{\mathcal{X}}=k_{\mathcal{A}} \otimes k_{\mathcal{Z}} = k_{\mathcal{A}}(a,a')\cdot k_{\mathcal{Z}}(z,z')$, where $k_{\mathcal{A}}(a,a') : \mathcal{A} \times \mathcal{A} \to \mathbb{R} $ measures how the cost changes when the PID parameters change, and $k_{\mathcal{Z}}(z,z') : \mathcal{Z} \times \mathcal{Z} \to \mathbb{R}$ measures how the cost change when the context changes. The intuition behind the multiplication of kernels is that we expect the cost to be similar if both PID parameters \emph{and} the outside air temperature are similar. 
For the PID parameters, we use the Mat\'ern 5/2 kernel to model similarities across the parameters' space and for the context the squared exponential kernel. We base this choice on prior knowledge about the smoothness of the costs function for both PID parameters and temperature. By using the Mat\'ern 5/2 kernel, we encode functions which are twice differentiable, and by using the squared exponential kernel, we model smooth (infinitely differentiable) functions. 

\paragraph{Kernel Hyperparameters.}
In GP regression, the kernel hyperparameters are usually estimated with a maximum likelihood estimation over the data. However, when data is actively acquired using the learned functions, optimizing the hyperparameters on these data leads to poor results \citep{bull2011convergence}. When this is done, the GP confidence interval is not guaranteed to contain the true function, and the optimization gets stuck in poor local optima. Thus, we first learn the hyperparameters via maximum likelihood by applying random safe inputs to the input space $\mathcal{A}$ while observing different contexts. During the optimization, we do not adapt the kernel hyperparameters with data but treat the hyperparameters as fixed, and the kernel defines fixed priors over functions.

\subsection{Comparison Baselines}
We use the following baselines to assess our algorithm performance. (1) The robust parameters originally used for PI room temperature control, which are currently deployed by Siemens Buildings Technology. (2) A Bayesian Optimization algorithm with no context nor safety constraints. (3) A Bayesian Optimization algorithm with outside air temperature as context, but no safety constraints. And (4) a model based variant where the parameters of Eq.~\ref{eq:first_order} are learned from data and optimal parameters are found using Ziegler-Nichols tuning rules for this idealized system, such as in \citep{seem1998new}. Furthermore, the model based version adapts the gains online, while the Bayesian Optimization based rules only change the gains once a day. We test all the benchmarks with the same real-life disturbance realizations as our algorithm.

\section{Experimental Results}
We use the simulator to evaluate the performance of the algorithm as if it was to be applied to the real room during a whole heating season. At the beginning of each day, we measure the outside air temperature, and we set the controller parameters according to Algorithm~\ref{alg:safe contextual ucb}. At the end of each day, we evaluate the cost function in Eq.~\ref{eq:cost function}. 

\subsection{Comparison Baselines}
\begin{figure}[ht]
	\begin{center}
	\includegraphics[width=\columnwidth]{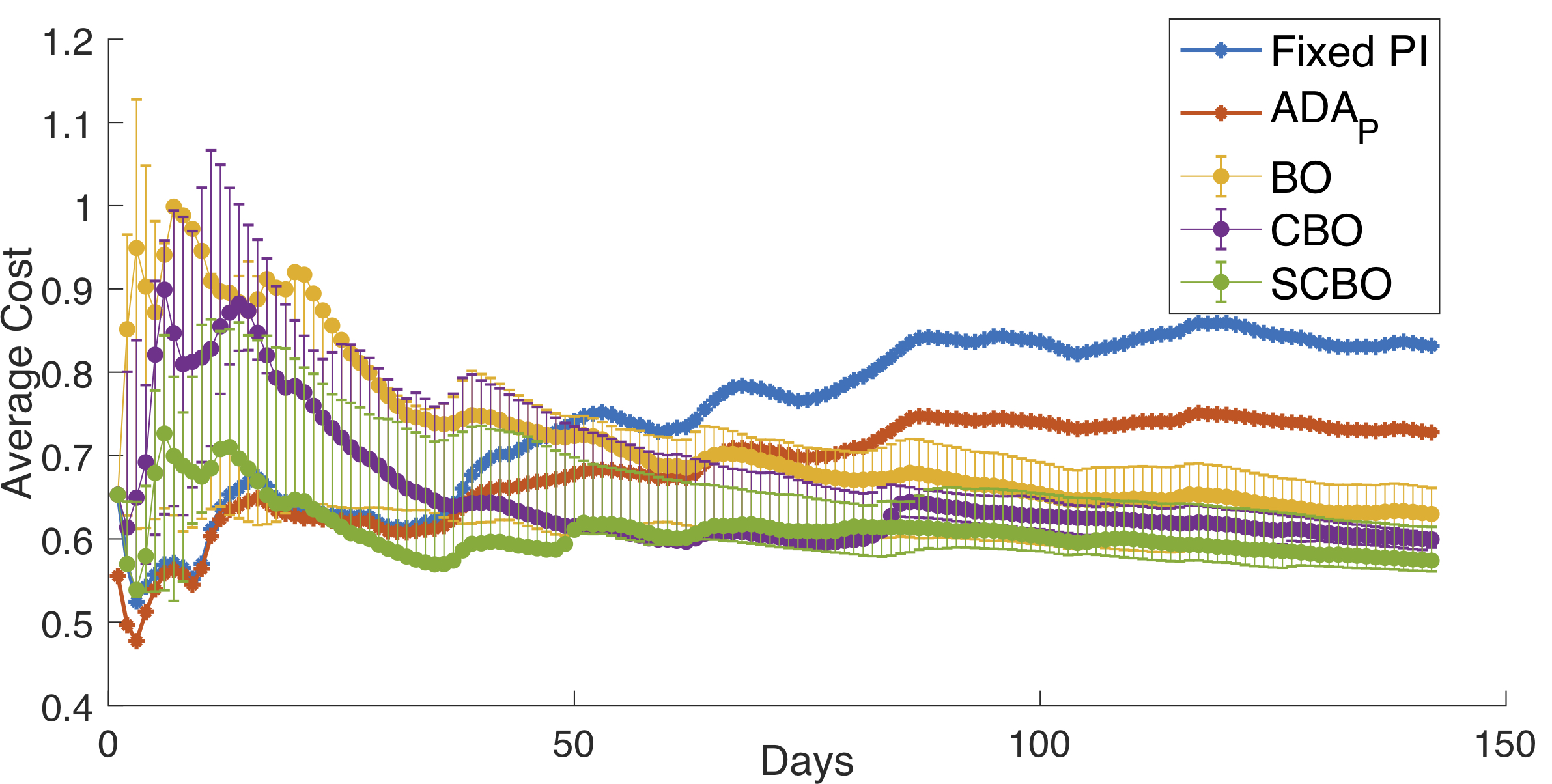}
	\caption{Cumulative average cost of the different algorithms. For each optimization algorithm, we plot the median of the average cost with filled markers, and the pointwise minimum and maximum with asymmetric error bars for five different random seeds. The Fixed PI is the robust PI controller, ADA$_\text{p}$ is the adaptive model-based baseline, BO is the vanilla Bayesian Optimization baseline, CBO is the Contextual Bayesian Optimization algorithm, and SCBO is the Safe Contextual Bayesian Optimization.}
	\label{fig:avg_cost_comp}
	\end{center} 
\end{figure}

In Fig.~\ref{fig:avg_cost_comp}, we plot the cumulative average cost for each day of the heating season. We initialize each algorithm with five different random seeds. We plot the median average cost with filled markers and the pointwise minimum and maximum average costs with asymmetric error bars. The reader can see that our algorithm, in green with SCBO legend, outperforms all the benchmarks, and we show the exact median values in Table~\ref{tab:average_reduction}. Constructively, we can see that the Bayesian Optimization variant already outperforms the fixed PID structure after half of the heating season (see the yellow line with legend BO). This is expected as it can freely optimize the gains. The next step is to add context to the Bayesian Optimization. This also outperforms the variant without context after half of the heating season (see the purple line with legend CBO). Nevertheless, this comes at the cost of large exploration peaks in the first iterations. The increase in the domain size requires more samples to learn the function. To limit the exploration to regions where the performance is not worse than the original controller, we introduce the safety constraints. In the first few iterations, our method performs as the fixed PID and, after ten days, it already outperforms all the benchmarks (see the green line with legend SCBO). As a final benchmark, we have the model based variant. This adaptation rule varies the controller online. Thus, after the first day, it performs better than all the other methods that start with the fixed PID gains (see the red line with legend ADA$_{\text{p}}$). Nevertheless, the system is not exactly first-order, and the Ziegler-Nichols rule is thus inaccurate, leading to worse performances. 

{\setlength{\tabcolsep}{7pt}
	\begin{table}[ht]
	\centering
	\begin{tabular}{cc}
		\hline \hline
		\multirow{2}{*}{Tuning Method} & Median \\ & Improvement\\
		\hline 
		Model Based Opt. (ADA$_{\text{p}}$) & 13\% \\ 
		Bayesian Opt. (BO) & 24\% \\
		Contextual Bayesian Opt. (CBO) & 28\% \\
		Safe-Contextual Bayesian Opt. (SCBO) & 32\% \\
		\hline \hline
	\end{tabular}
	\caption{Median of the average cost reduction of the different algorithms after a heating season with respect to the fixed PI controller.}
	\label{tab:average_reduction}
	\end{table}
}

\subsection{Safe Set Expansion}
\begin{figure}[ht]
	\begin{center}
	\includegraphics[width=0.4\columnwidth]{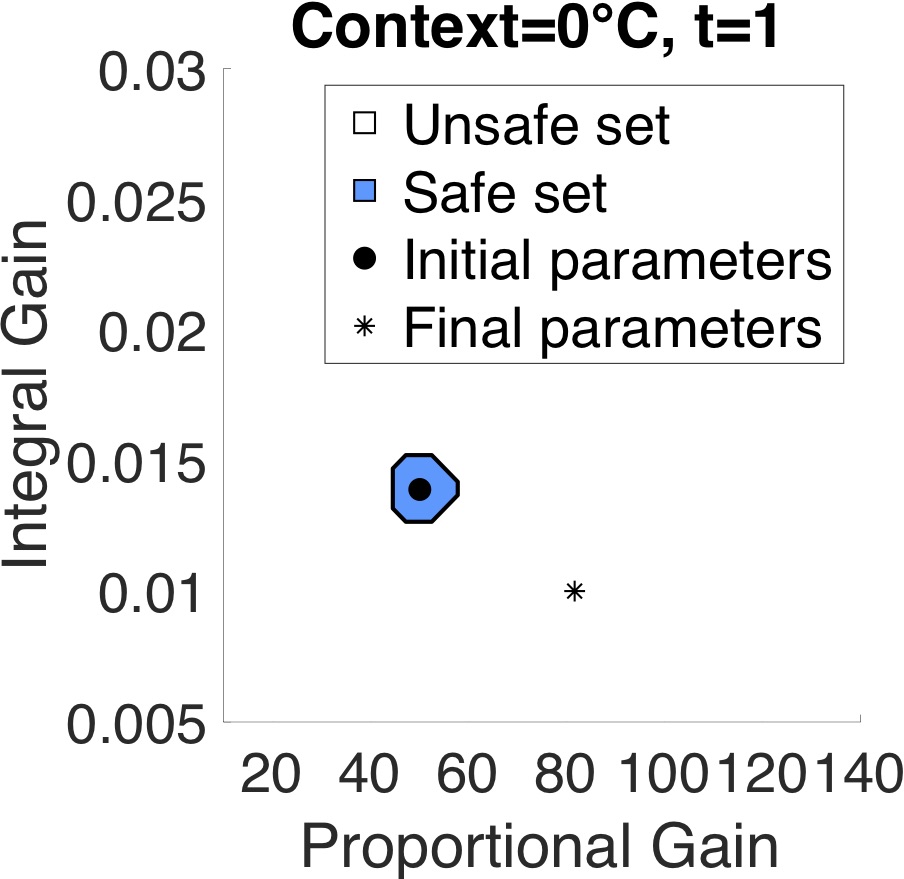}
	\includegraphics[width=0.4\columnwidth]{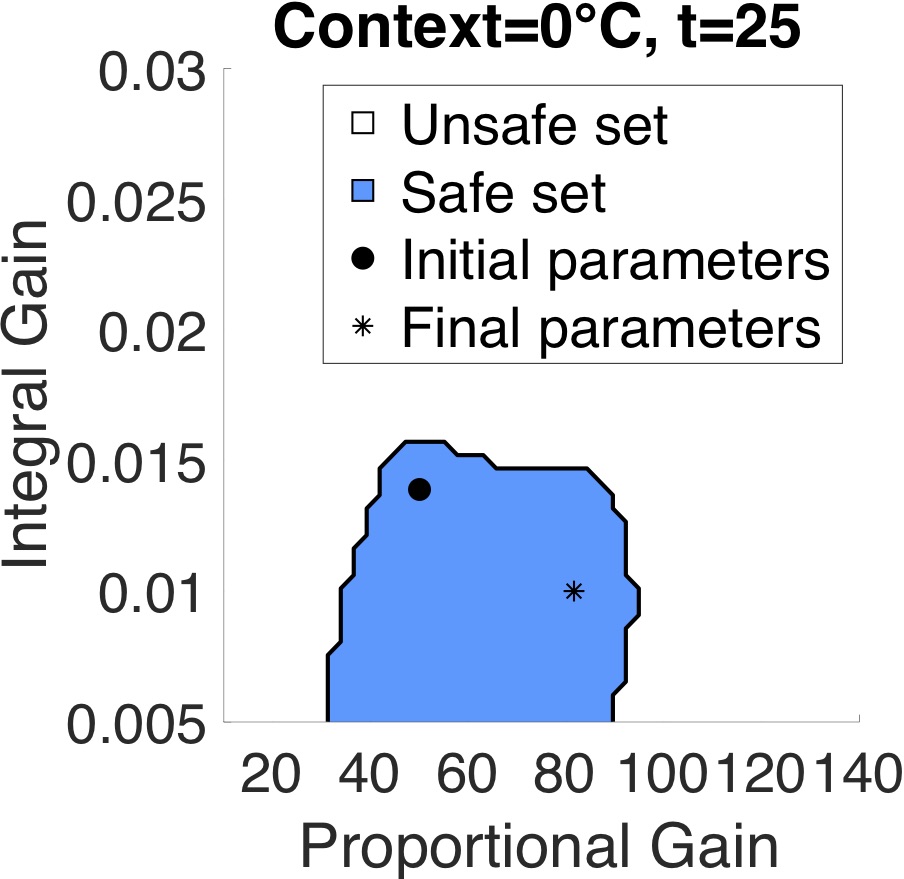}\\
	\includegraphics[width=0.4\columnwidth]{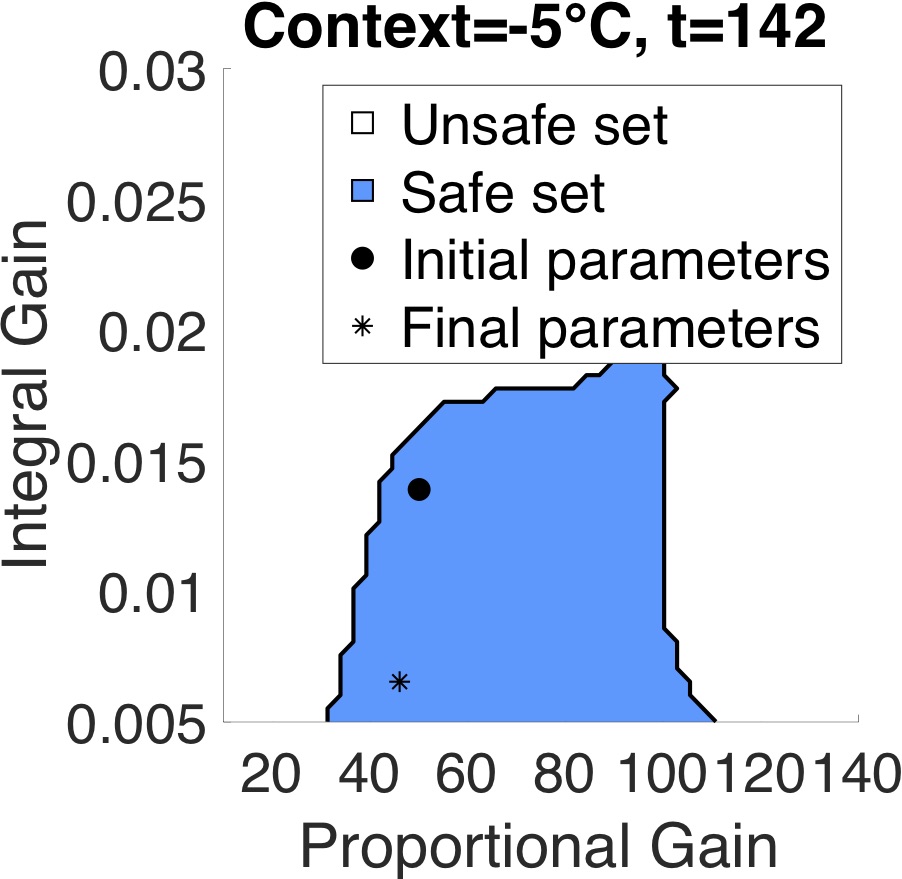}
	\includegraphics[width=0.4\columnwidth]{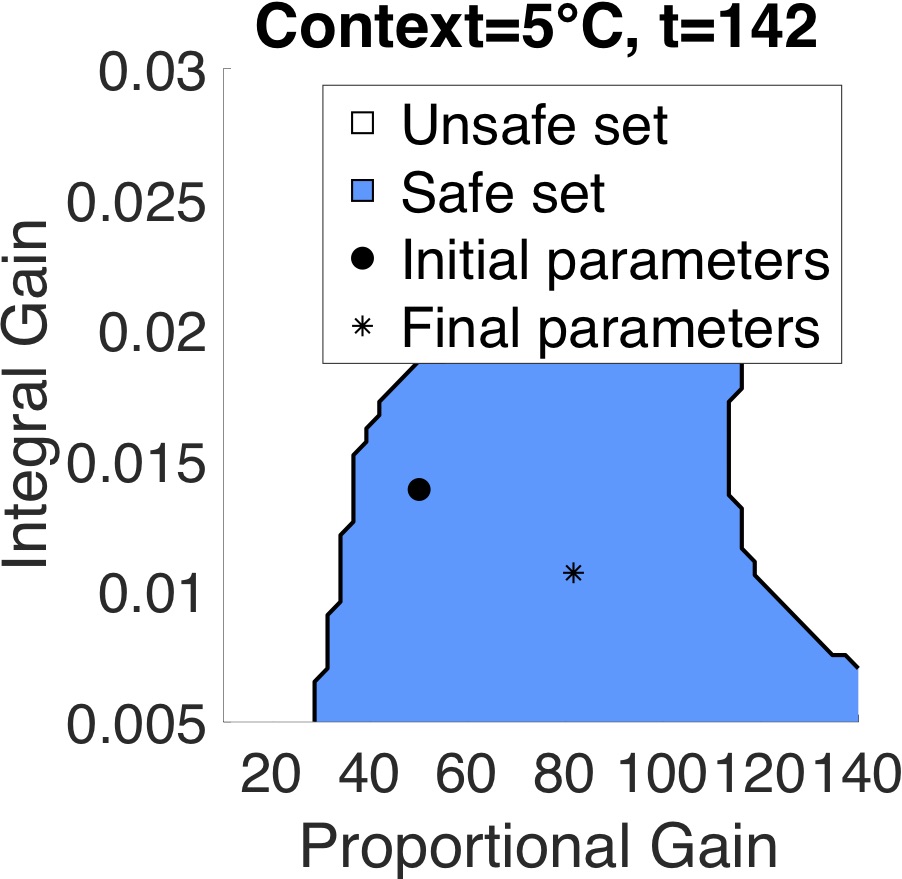}
	\caption{Parameters safe set in light blue and unsafe set in white. The top row shows the safe set when the context is $0^\circ C$ and time is 0 and 25 days. The bottom row shows the final safe set for $-5^\circ C$ and $+5^\circ C$.
	The initial PI parameters are shown with a black circle while the final optimized parameters with a black star.}
	\label{fig:safe_dom_time}
	\end{center} 
\end{figure}

In Fig.~\ref{fig:safe_dom_time}, we plot the safe set in light blue at different times and contexts, while in white, we show the unsafe region. Initially, in the top-left subfigure, only a small set around the initial parameters is considered safe. As shown in the top row of Fig.~\ref{fig:safe_dom_time}, the safe set expands with the following mechanism: as the algorithm queries points in the boundary, its uncertainty in the neighborhood of these points decreases. As it becomes more confident of points with high performance, it includes these points in the set $S_n$. For example, the optimal point is outside the initial safe set but, after the $25^\text{th}$ iteration, is contained in the safe set. Finally, even if our algorithm does not expand the safe set explicitly with the acquisition function as does SafeOPT \citep{sui2015safe}, we hypothesize that this is generated by the landscape of the optimization problem.% in Eq.~\ref{eq:minimization}.

If the Contextual Bayesian Optimization (CBO) without safety variant were to be used, the exploration would have covered the whole PI space to learn that a big part of the domain performs worse than the fixed PI. This is shown in the exploratory peaks at the first iterations of the CBO plot in Fig.~\ref{fig:avg_cost_comp}. Nevertheless, the global optimum is inside the safe region. Hence, CBO and SCBO find the same optimizers. This explains why the performance curve in Fig.~\ref{fig:avg_cost_comp} tends to the Safe Contextual Bayesian Optimization algorithm. 

In the bottom row of Fig.~\ref{fig:safe_dom_time}, we show how the safe set changes with context at the end of the heating season. This shows that context affects the underlying process and, if not included, it could lead to unsafe exploration. For example, imagine that the learning is done only during $T=+5^\circ$ because of an environmental scenario with a Safe Bayesian Optimization algorithm without context. If the temperature changes to $T=-5^\circ$, then there is no notion of uncertainty at this new temperature and an unsafe point could be used. 

\subsection{Minimizers Gain Scheduling}
\begin{figure}[hb]
	\begin{center} 
	\includegraphics[width=\columnwidth]{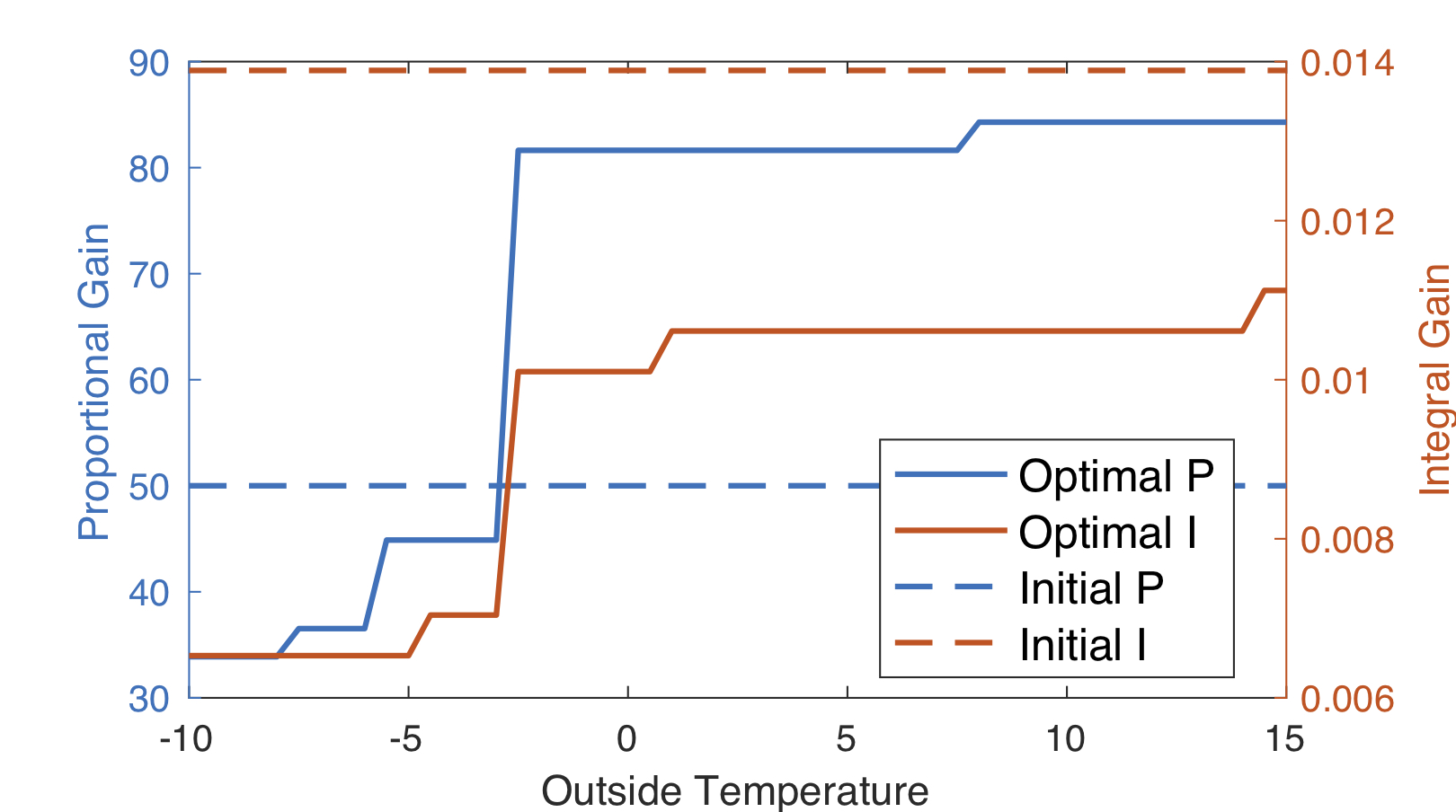}
	\caption{Optimal controller parameters as a function of the outside temperature. In full lines, we plot the optimal proportional and integral gains and in dashed lines, the initial controller parameters. This plot can be used as a gain scheduler after the learning phase.} 
	\label{fig:Contourcomp}
	\end{center} 
\end{figure}

In Fig.~\ref{fig:Contourcomp} we plot how the optimal PI parameters change with the outside air temperature at the end of the heating season. This plot can be used after the learning phase as a gain scheduler. The significant change at -3 degrees is due to the interaction with the automatic weather compensation that changes the water temperature. Hence, actuation power decreases, and the gains increase considerably. This proves that including context and capacity to adapt to it is critical for performance. 

\newpage
\section{Conclusions}
In this work, we optimized the PI controller parameters in a ubiquitous room temperature control loop. The proposed cost function is very simple and interpretable to the practitioner, but the exact relationship with the controller parameters are complex and unknown. We used a Bayesian Optimization algorithm that explicitly trades-off exploitation and safe exploration to learn and optimize the cost function simultaneously. Furthermore, it includes a context to adapt to varying environmental conditions. In experiments, we showed that our algorithm outperforms the baselines and paves the way to optimizing other control loops in HVAC systems.

% % \section*{Acknowledgments}
% % We would like to thank the Building Technologies division of Siemens Schweiz AG for providing the plant model and the disturbances measurements used in this research.

\newpage
\bibliographystyle{named}
\bibliography{refs}

\appendix

\end{document}